\documentclass[11pt]{article}

\usepackage[utf8]{inputenc}
\usepackage[T1]{fontenc}
\usepackage{amsmath,amssymb,amsfonts}
\usepackage{graphicx}
\usepackage{booktabs}
\usepackage{multirow}
\usepackage{xcolor}
\usepackage{hyperref}
\usepackage{cleveref}
\usepackage[margin=1in]{geometry}
\usepackage{subcaption}
\usepackage{enumitem}
\usepackage{url}
\usepackage{tikz}
\usetikzlibrary{arrows.meta,positioning,calc,fit,backgrounds}

\hypersetup{
    colorlinks=true,
    linkcolor=blue!60!black,
    citecolor=green!50!black,
    urlcolor=blue!70!black,
}

\newcommand{\loo}{\text{LOO}}
\newcommand{\deltaloo}{\Delta_{\loo}}

\title{Structural Evaluation Metrics for SVG Generation\\via Leave-One-Out Analysis%
\thanks{Demo dashboard: \url{https://haonan-lica.github.io/svg-structural-metrics-dashboard/}}}

\author{
    Haonan Zhu \hspace{0.7em} Adrienne Deganutti \hspace{0.7em} Elad Hirsch\hspace{0.7em} Purvanshi Mehta \\
  \texttt{\{haonan, adrienne, elad, purvanshi\}@lica.world} \\
}

\date{}

\begin{document}
\maketitle

\begin{abstract}
SVG generation is typically evaluated by comparing rendered outputs to reference images, which captures visual similarity but not the structural properties that make SVG editable, decomposable, and reusable. Inspired by the classical jackknife~\cite{quenouille1956,tukey1958}, we introduce \textbf{element-level leave-one-out (LOO) analysis}. The procedure renders the SVG with and without each element which yields element-level signals for quality assessment and structural analysis. From this single mechanism we derive (i)~per-element quality scores that enable zero-shot artifact detection (F1~$\geq$~0.87, $+0.17$ over baselines); (ii)~element--concept attribution via LOO footprints crossed with VLM-grounded concept heatmaps; and (iii)~four structural metrics: \emph{purity}, \emph{coverage}, \emph{compactness}, \emph{locality} that quantify SVG modularity from complementary angles. These metrics extend SVG evaluation from image similarity to code structure, enabling element-level diagnosis and comparison of how visual concepts are represented, partitioned, and organized within SVG code. Their practical relevance is validated on over 19,000 edits (5 types) across 5 generation systems and 3 complexity tiers.
\end{abstract}

\section{Introduction}
\label{sec:intro}

Scalable Vector Graphics (SVG) describe images as geometric primitives (paths, shapes, text) in XML. Unlike raster generation, SVG generation produces \emph{structured code} that can be inspected, edited, and version-controlled. While recent work formulates SVG generation as a code-completion problem for autoregressive language models~\cite{starvector,omnisvg}, evaluation remains confined to the rendered raster image space, where generated outputs are compared with reference images using metrics such as CLIP~\cite{radford2021clip} or FID.

This single-score evaluation has three blind spots:
\begin{enumerate}[leftmargin=*,itemsep=2pt]
\item \textbf{No element-level diagnosis.}  A final score cannot tell
whether every element contributes positively or whether one bad element
ruins an otherwise good SVG.
\item \textbf{No concept grounding.}  We cannot tell which visual
concept (``the flower'' vs.\ ``the stem'') each element serves.
\item \textbf{No structural assessment.}  Two SVGs can render identically
yet differ in modularity: one uses separate elements per concept, another
packs everything into a single compound path.
\end{enumerate}

\noindent We address all three through \textbf{leave-one-out (LOO) analysis},
grounded in the classical jackknife
principle~\cite{quenouille1956,tukey1958}: for each element $e_i$, render
the SVG with and without $e_i$ and compare the two images.  Each pair of
renders produces two complementary signals.  First, a scalar quality score
measures the element's global contribution:
\begin{equation}
    \deltaloo(e_i) = S(\text{SVG}) - S(\text{SVG} \setminus \{e_i\})
    \label{eq:loo}
\end{equation}
where $S$ is a reference-based similarity (CLIP in our experiments).  A
positive $\deltaloo$ indicates a helpful element; a negative value means
removing it would improve the image.  Second, a pixel-level difference map per pixel index $x$ and $y$:
\begin{equation}
    M_i(x,y) = \bigl|I(\text{SVG})(x,y) - I(\text{SVG} \setminus \{e_i\})(x,y)\bigr|
    \label{eq:pixeldiff_intro}
\end{equation}
localizes the spatial footprint of each element, enabling concept
attribution (\Cref{sec:attribution}).  Both signals are extracted from the
same set of $N$ LOO renders at no additional cost.

\begin{figure}[t]
\centering
\includegraphics[width=0.9\textwidth]{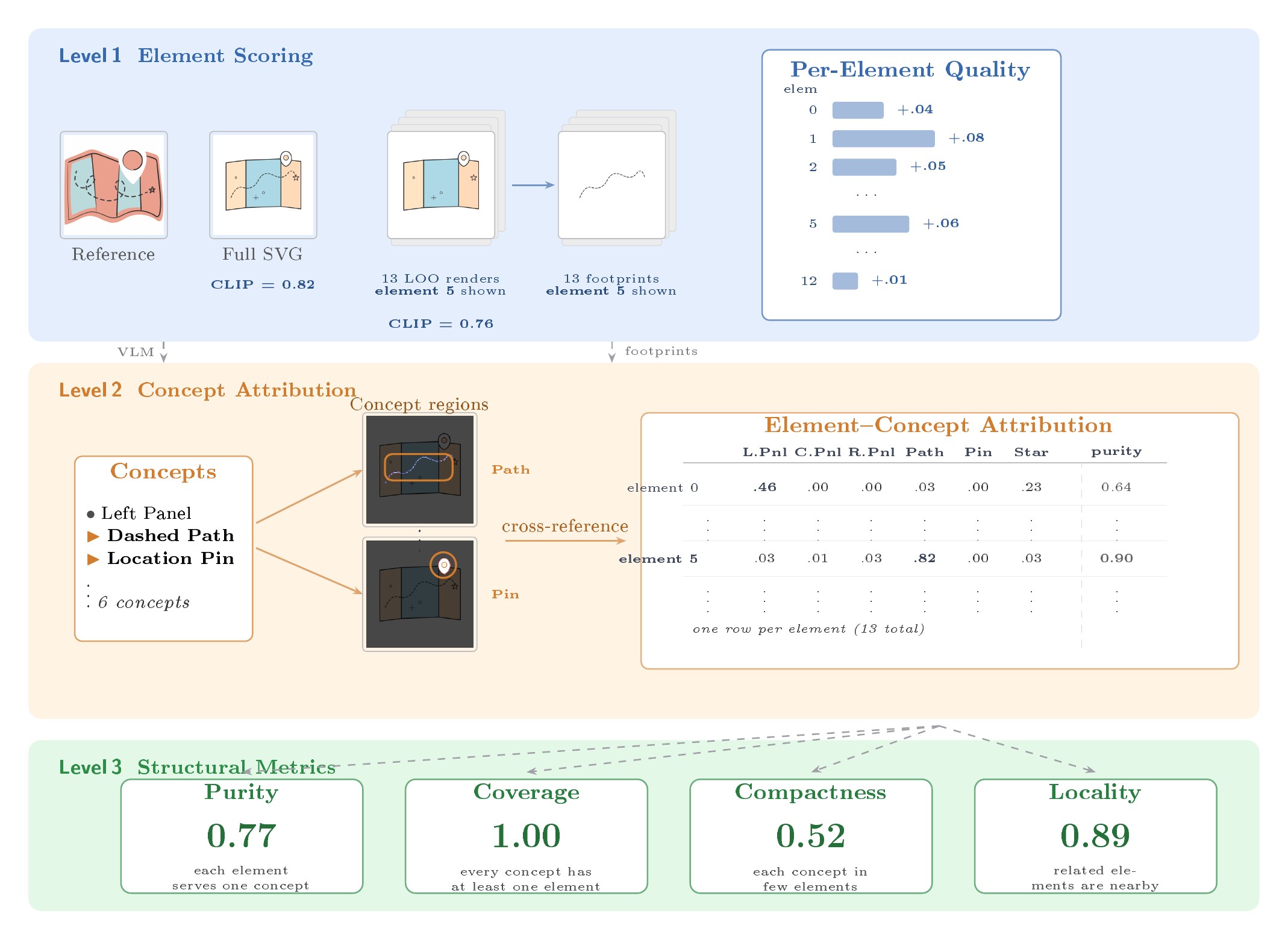}
\caption{Framework overview on a 13-element map SVG.
Level 1 (top): comparing the full render with each of $N$
leave-one-out renders yields per-element CLIP-delta scores and
pixel-difference footprints.
Level 2 (middle): a VLM lists visual concepts, CLIPSeg and SAM3 ground
each spatially, and crossing footprints with concept heatmaps gives an
$N{\times}C$ attribution matrix whose row-max is the element's purity.
Level 3 (bottom): four metrics aggregate the matrix, asking whether
each element is dedicated to one concept (purity), each concept has a
dedicated element (coverage), a concept concentrates in few elements
(compactness), and a concept's elements sit near one another in
z-order (locality).}
\label{fig:overview}
\end{figure}

Just as the James--Stein estimator~\cite{james1961} showed that simple shrinkage can provide better statistical guarantees, our LOO decomposition extracts rich structural information while only requires re-rendering. Unlike SVG generation methods that perturb element ordering during training~\cite{neuralsvg,t2vnpr}, LOO preserves the z-index of every remaining element in each ablation, ensuring that occlusion relationships are maintained. See Fig. (\Cref{fig:overview}) for an overview of the proposed framework. 

In summary, our main contributions are the following: 
\begin{enumerate}[leftmargin=*,itemsep=2pt]
\item \textbf{Element scoring} (\Cref{sec:scoring}): per-element quality
signals that identify harmful elements and enable artifact removal.
\item \textbf{Concept attribution} (\Cref{sec:attribution}): LOO
pixel-difference masks crossed with concept heatmaps produce an
element--concept attribution matrix.
\item \textbf{Structural metrics} (\Cref{sec:metrics}): four metrics
derived from the attribution matrix (purity, coverage, compactness, and
locality) that characterize SVG modularity from complementary angles.
This is our primary contribution.
\end{enumerate}

\section{Related Work}
\label{sec:related}

\paragraph{SVG generation.}
Neural SVG generation has progressed from stroke-level RNNs~\cite{sketchrnn}
and variational autoencoders~\cite{deepsvg} to autoregressive code
models~\cite{starvector,omnisvg} that generate full SVG markup from text
or image prompts.  Evaluation relies on rendered-image metrics (CLIP, FID,
human preference), none of which assess code structure.

\paragraph{Element ordering and sampling in SVG generation.}
SVG rendering is order-dependent: later elements occlude earlier ones.
Several generation methods perturb this ordering during training.
NeuralSVG~\cite{neuralsvg} applies nested dropout, randomly truncating
the path sequence to encourage meaningful layering.
Neural Path Representations~\cite{t2vnpr} re-sort paths by area for
coarse-to-fine optimization.  DeepSVG~\cite{deepsvg} explores
lexicographic and Hungarian assignment to handle the path-ordering
ambiguity.  These strategies are designed for generation, not evaluation;
applying them to measure element contributions would conflate ordering
effects with element quality.  Our LOO approach removes exactly one element
at a time while preserving the order of all remaining elements, mirroring
the jackknife estimator~\cite{quenouille1956,tukey1958} whose statistical
guarantees rely on minimal, structured perturbations.

\paragraph{Per-step evaluation.}
Process reward models provide step-level signals for math
reasoning~\cite{lightman2023lets} and code
generation~\cite{uesato2022solving,chen2021codex}.  Our LOO scoring serves
a similar role (per-element quality) but is derived from rendering rather
than learned annotations.

\paragraph{Disentangled representations.}
The DCI framework~\cite{eastwood2018dci} evaluates learned representations
along three axes: disentanglement (each code captures one factor),
completeness (each factor is captured), and informativeness.  We extend
this framework to SVG structure, where ``codes'' are XML elements,
``factors'' are visual concepts, and the importance matrix is constructed
from LOO rendering rather than learned weights.

\paragraph{SVG editability.}
Editability has been studied for radiance fields and diffusion models ~\cite{instructnerf2023}, where it is considered a property of the learned latent representation. In vector graphics, editability instead depends on the code structure itself, e.g., whether visual concepts map cleanly to the XML elements.

\section{Method}
\label{sec:method}

\subsection{Element-Level LOO Scoring}
\label{sec:scoring}

Given an SVG with $N$ visual elements $\{e_1, \ldots, e_N\}$, we compute
the LOO delta (\Cref{eq:loo}) for each element using CLIP ViT-B/32
image-to-image similarity at $384 \times 384$ pixels, rendered via CairoSVG~\cite{cairosvg}.  This requires
$N$ renders, each with one of $N$ elements removed, plus one complete render.  We classify
elements as \emph{helpful} ($\deltaloo > 0.005$), \emph{harmful}
($\deltaloo < -0.005$), or \emph{neutral}, following the terminology of
Koh and Liang~\cite{koh2017influence} for leave-one-out influence.

\paragraph{Subpath splitting.}
Many SVGs contain a single \texttt{<path>}\footnote{\texttt{<path>} are sequence of drawing commands that are often used to create complex shapes.} element encoding multiple disjoint shapes via repeated \texttt{M} (moveto) commands.  We split compound paths at these boundaries, increasing the median scoring units from 2 to 9 across our validation set.

\subsection{Concept Attribution}
\label{sec:attribution}

LOO scoring quantifies each element’s contribution to overall visual quality, but does not identify which visual concept the element corresponds to. We bridge this gap in following three steps:

\paragraph{Step 1: Concept extraction.}
A vision-language model (Qwen3-VL-32B) examines the rendered SVG and lists distinct visual concepts (\textit{e.g.}, ``red flower head'', ``green stem'').  

\paragraph{Step 2: Concept grounding.}
For each concept $c_j$, we produce a spatial heatmap $H_{c_j}(x,y) \in [0,1]$ indicating which pixels belong to it.  We use a two-model fusion strategy.  First, CLIPSeg~\cite{clipseg} (\texttt{clipseg-rd64-refined}) takes the rendered image and each concept as a text prompt, producing sigmoid-normalized soft heatmaps at $352 \times 352$ resolution (bilinearly upsampled to the render size).  CLIPSeg inherits CLIP's open vocabulary, making it effective for abstract or compositional concepts (e.g., ``ornate border''). Second, when pre-computed SAM3 masks are available, we prefer them for concepts where SAM3 returns a confident detection (score $\geq 0.3$, area fraction in $[0.005, 0.95]$), since SAM3 provides pixel-precise binary boundaries for concrete objects.  Per concept, the pipeline selects the SAM3 mask if it passes these quality filters, otherwise falls back to the CLIPSeg heatmap. Concepts whose grounding masks overlap heavily (IoU $> 0.9$) are merged.

\begin{figure}[htbp]
\centering
\includegraphics[width=\textwidth]{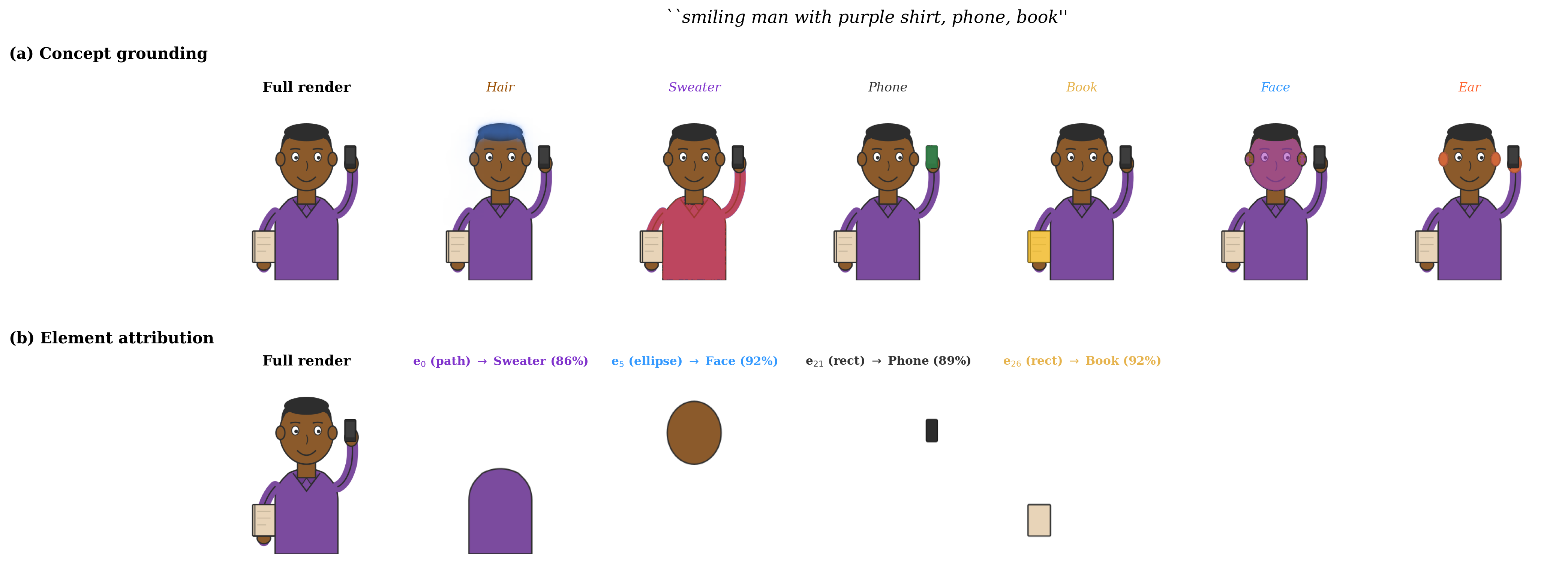}
\caption{Concept attribution example (Claude, complex tier).
\textbf{(a)}~Concept grounding: six concepts are automatically extracted
and spatially localized; each panel highlights one concept region while
dimming the rest.
\textbf{(b)}~Element attribution: four elements rendered in isolation,
each labeled with its primary concept and attribution share from
attribution matrix $\mathbf{A}$ (defined in Eq. \ref{eq:attribution}). The sweater path maps to ``Sweater'' (86\%), the phone
rect to ``Phone'' (89\%), the book rect to ``Book'' (92\%), and the face
ellipse to ``Face'' (92\%).}
\label{fig:concept_example}
\end{figure}

\paragraph{Step 3: Element concept attribution.}
The LOO pixel-difference mask for element $e_i$ is:
\begin{equation}
    M_i(x,y) = \bigl| I(\text{SVG})(x,y) -
    I(\text{SVG} \setminus \{e_i\})(x,y) \bigr|
    \label{eq:pixeldiff}
\end{equation}
The concept contribution of $e_i$ to $c_j$ is the normalized overlap:
\begin{equation}
    A(e_i, c_j) = \frac{\sum_{x,y} M_i(x,y) \cdot H_{c_j}(x,y)}
                       {\sum_{x,y} M_i(x,y) + \epsilon}
    \label{eq:attribution}
\end{equation}
This yields an attribution matrix
$\mathbf{A} \in \mathbb{R}^{N \times C}$.  From $\mathbf{A}$ we assign
each element a primary concept
$c^*(e_i) = \arg\max_j A(e_i, c_j)$ and a purity score:
\begin{equation}
    \text{purity}(e_i) = \frac{\max_j A(e_i, c_j)}
                              {\sum_j A(e_i, c_j) + \epsilon}
    \label{eq:purity}
\end{equation}
Elements with total attribution below 0.01 are considered \emph{inactive}
(negligible visual footprint) and excluded from metric computations. Fig. \ref{fig:concept_example} illustrates this through an example, where given an SVG of a person holding a phone and book, we automatically identify six spatial concepts and map each SVG element to its primary concept. 

\subsection{Structural Metrics}
\label{sec:metrics}

The attribution matrix $\mathbf{A}$ enables us to ask: \emph{can each
visual concept be independently modified?}  We define four metrics,
extending the DCI framework~\cite{eastwood2018dci} from continuous latent
representations to discrete SVG elements.  Each metric captures a different
structural property relevant to editing.

\paragraph{Purity (disentanglement).}
Mean purity measures whether each element serves a single concept:
\begin{equation}
    \overline{\text{purity}} = \frac{1}{|\mathcal{A}|}
    \sum_{e_i \in \mathcal{A}} \text{purity}(e_i)
\end{equation}
where $\mathcal{A}$ is the set of active elements.  Purity of 1.0 means
every element is dedicated to one concept; low purity means editing an
element affects multiple concepts.

\paragraph{Coverage (completeness).}
The fraction of visual concepts with at least one dedicated element:
\begin{equation}
    \text{coverage} = |\{c_j : |G(c_j)| \geq 1\}| \;/\; C
\end{equation}
where $G(c_j) = \{e_i : c^*(e_i) = c_j\}$ is the concept group.
Coverage below 1.0 means some concepts cannot be individually addressed.

\paragraph{Compactness.}
Whether each concept is represented by few elements (easy to locate) or
fragmented across many, measured by the normalized Herfindahl index:
\begin{equation}
    \text{compactness}(c_j) = \frac{H(c_j) - 1/n_j}{1 - 1/n_j},
    \quad H(c_j) = \sum_i \left(\frac{A(e_i,c_j)}
    {\textstyle\sum_k A(e_k,c_j)}\right)^{\!2}
\end{equation}
where $n_j$ is the number of active elements for concept $c_j$.
Compactness of 1.0 means a single element captures the concept.

\paragraph{Locality.}
How close together a concept's elements are in the SVG source order
(z-order), weighted by attribution.  We compute the attribution-weighted
mean absolute deviation from the centroid, inspired by the Earth Mover's
Distance:
\begin{equation}
    \text{locality}(c_j) = 1 -
    \frac{\sum_i w_i \, |i - \mu_j|}{(N{-}1)/2}
\end{equation}
where $w_i = A(e_i, c_j) / \sum_k A(e_k, c_j)$ and
$\mu_j = \sum_i w_i \cdot i$.  Unlike binary adjacency, locality is
continuous: elements at positions [1, 3, 5] score higher than [1, 9, 18],
and high-attribution elements far from the centroid are penalized more.

\medskip
Together, these four metrics characterize distinct failure modes in code quality.: purity $\to$  leaks across concepts; coverage $\to$ some concepts are not addressable; compactness $\to$ a concept is fragmented across many elements; locality $\to$ a concept's elements are scattered in the file.

\section{Experimental Setup}
\label{sec:setup}

\paragraph{Dataset.}
300 validation SVGs are bootstrapped from the LICA~\cite{hirsch2026lica} collection of layered graphic-design compositions and stratified into three complexity levels (100 each): \emph{simple}, \emph{medium}, and \emph{complex}, using the weighted complexity measure from~\cite{chen2025svgenius}. This stratification allows us to examine metric behavior across levels of complexity, from simple flat icons to multi-element compositions with substantial structural variation across model outputs. Each SVG has a text description and a reference rendering at $384 \times 384$.  Source SVGs are the original dataset files, filtered for renderability.

\paragraph{Models.}
We evaluate 5 SVG generation systems in following categories:
\begin{itemize}[leftmargin=0.1em,label={},itemsep=1pt]
\item \emph{Proprietary Models}: Claude~4.5-Opus, GPT-5.2, Gemini~3-flash-preview
\item \emph{Open Source Model}: Qwen3-Coder-30B-A3B-Instruct~\cite{qwen3coder} 
\item \emph{Vectorization}: VTracer \cite{vtracer} (deterministic rule based image-to-SVG)
\end{itemize}

\paragraph{Concept pipeline.}
Concepts are extracted by Qwen3-VL-32B and grounded via CLIPSeg and
SAM3 discussed in Section. \ref{sec:attribution}.  CLIP ViT-B/32 is used as the similarity backbone throughout.

\paragraph{Empirical validation.}
To assess whether structural metrics capture code quality, we test whether they predict downstream editing success\footnote{Ease of editing is one important property of high-quality code}. We perform five types of edits on each model's SVGs (\Cref{tab:edit_types})
and measure \textbf{edit precision}:
\begin{equation}
    \text{precision} = \frac{\text{target change}}
    {\text{target change} + \text{collateral damage}}
    \label{eq:edit_precision}
\end{equation}
where \emph{target change} is the pixel difference within the edited
concept's mask and \emph{collateral damage} is the difference in other
concepts' masks.  Masks are derived from the same concept grounding
pipeline (CLIPSeg + SAM3).  We correlate each metric with edit precision
using a train and test split across tiers, correcting for 5 comparisons
(Bonferroni, $\alpha = 0.01$).

\begin{table}[t]
\centering\small
\caption{Five edit operations used for empirical validation.}
\label{tab:edit_types}
\begin{tabular}{lll}
\toprule
Edit & Operation & Structural property tested \\
\midrule
Color  & Change fill or stroke to random color & Element--concept isolation \\
Delete & Remove all elements of a concept & Concept group completeness \\
Move   & Translate elements by 20\,px & Spatial independence \\
Scale  & Scale elements by $0.7\times$ & Spatial independence \\
Regroup & Reorder elements to be contiguous & Z-order locality \\
\bottomrule
\end{tabular}
\end{table}

\section{Results}
\label{sec:results}

\subsection{Artifact Detection}
\label{sec:artifact_results}

We inject 3 synthetic artifacts per SVG (random shapes, stray paths,
duplicated-with-offset elements) into clean reference SVGs and compare
six detection methods, each flagging exactly $K{=}3$ elements.

\begin{figure}[t]
\centering
\includegraphics[width=0.6\textwidth]{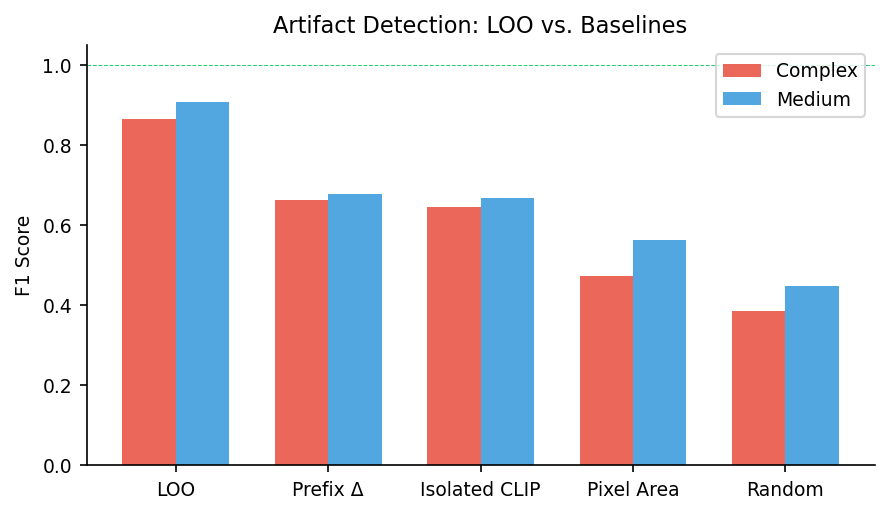}
\caption{Artifact detection F1 scores.  LOO outperforms all baselines by
$\geq$0.17 F1 and is the only method where removing flagged elements
improves SSIM ($\geq$+0.028).  The advantage comes from \emph{context}:
LOO evaluates each element within the full composition.}
\label{fig:artifact}
\end{figure}

\noindent We apply element-level LOO scores in Section \ref{sec:scoring} with threshold ($\deltaloo < -0.005$, \emph{harmful})) to detect these artifacts. Fig. \ref{fig:artifact} summarize the results. LOO achieves F1 $\geq$ 0.87, compared to $\leq$0.68 for the next-best methods (prefix delta, isolated CLIP).  The key difference is context: LOO measures each element's contribution within the full SVG, while baselines score elements independently or depend on ordering.  Critically, LOO is the only method that \emph{improves} visual quality when flagged elements are removed (positive $\Delta$SSIM), while baselines often make SVGs worse.

\subsection{Structural Metrics}
\label{sec:metric_results}

\paragraph{Model profiles.}
\Cref{tab:metrics} shows all four metrics for the complex tier.
Coverage is near-saturated ($>$0.96), as all generators produce elements
covering the requested concepts.  Purity is the most discriminative,
separating Claude/vtracer ($\geq$0.68) from Source SVG (0.60).  Locality
ranks models differently: vtracer leads (0.87) because its vectorization
traces connected regions, keeping related elements adjacent.  Source SVG
has high compactness (0.47) but the lowest purity, since compound paths are
concentrated but entangled.

\begin{table}[t]
\centering\small
\caption{Structural metrics, complex tier. Higher = more modular, except crosstalk (lower = better).}
\label{tab:metrics}
\begin{tabular}{lrrrrrr}
\toprule
Model & Purity & Cover. & Compact. & Locality & Crosstalk & Elem. \\
\midrule
Claude  & \textbf{0.70} & 0.98 & 0.33 & 0.76 & 0.32 & 27 \\
vtracer & 0.68 & 0.98 & 0.41 & \textbf{0.87} & \textbf{0.29} & 53 \\
Gemini  & 0.67 & 0.98 & \textbf{0.48} & 0.78 & 0.31 & 14 \\
GPT-4o  & 0.66 & 0.97 & 0.35 & 0.75 & 0.35 & 28 \\
Qwen3-Coder & 0.64 & 0.97 & 0.44 & 0.74 & 0.37 & 17 \\
Source SVG & 0.60 & 0.98 & 0.47 & 0.80 & 0.35 & 46 \\
\bottomrule
\end{tabular}
\end{table}

\begin{figure}[t]
\centering
\includegraphics[width=\textwidth]{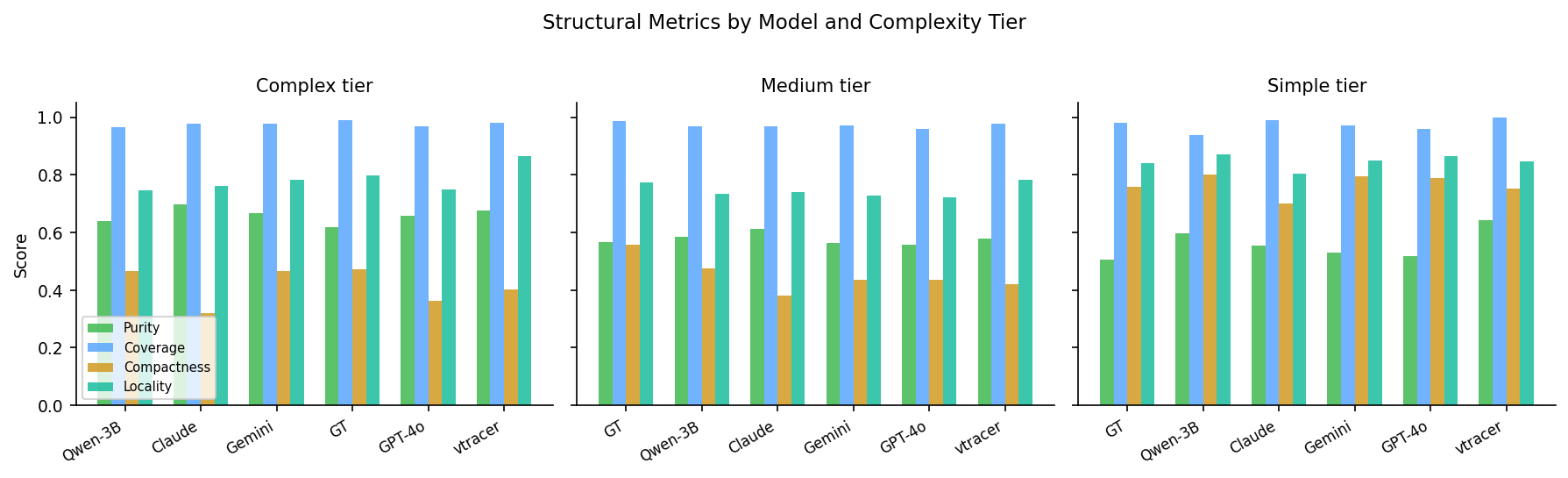}
\caption{Structural metrics across all three complexity tiers.  Purity
shows the largest between-model variance; coverage is near-saturated.
Locality reveals a distinct ranking (vtracer leads) compared to purity
(Claude leads).}
\label{fig:metrics}
\end{figure}

\paragraph{Empirical validation.}
\Cref{tab:empirical} shows edit precision for the complex tier.  Model
rankings from empirical edits closely match the purity ranking.

\begin{table}[t]
\centering\small
\caption{Empirical edit precision, complex tier (${\sim}$1{,}500 edits per
model).  Higher = edits better localized to the target concept.}
\label{tab:empirical}
\begin{tabular}{lrrrrrr}
\toprule
Model & Overall & Color & Delete & Move & Scale & Regroup \\
\midrule
Claude  & \textbf{0.80} & 0.85 & \textbf{0.88} & \textbf{0.83} & \textbf{0.86} & 0.61 \\
vtracer & 0.80 & \textbf{0.86} & 0.87 & 0.79 & 0.83 & \textbf{0.66} \\
Gemini  & 0.79 & 0.83 & 0.87 & 0.82 & 0.84 & 0.58 \\
GPT-4o  & 0.76 & 0.78 & 0.84 & 0.78 & 0.80 & 0.59 \\
Qwen3-Coder & 0.74 & 0.80 & 0.82 & 0.75 & 0.79 & 0.56 \\
Source SVG & 0.56 & 0.58 & 0.58 & 0.56 & 0.57 & 0.51 \\
\bottomrule
\end{tabular}
\end{table}

\begin{figure}[t]
\centering
\includegraphics[width=\textwidth]{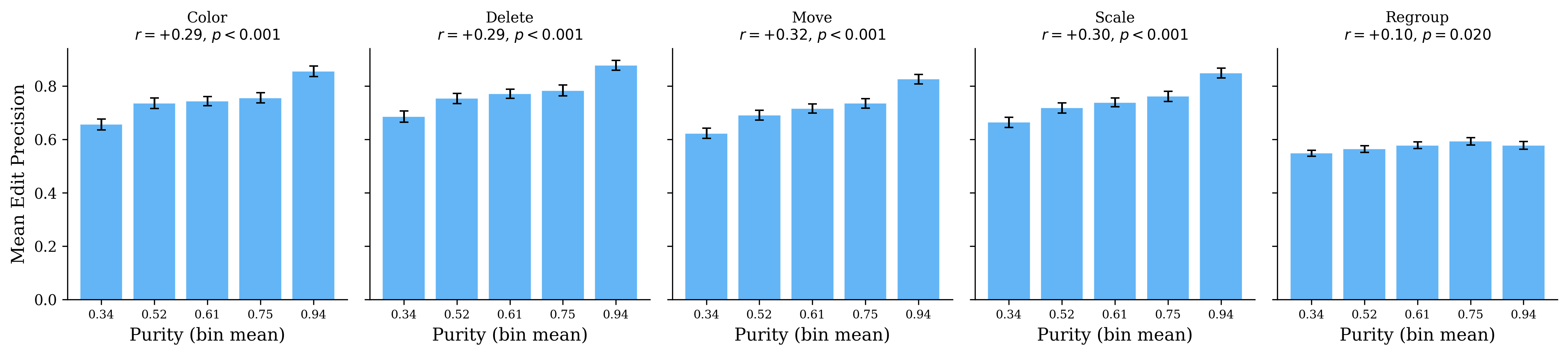}
\caption{SVG-level purity vs.\ edit precision, binned into quintiles
($n \approx 120$ SVGs per bin).  Higher purity SVGs are consistently easier
to edit across all five edit types.}
\label{fig:binned}
\end{figure}

\begin{figure}[t]
\centering
\includegraphics[width=\textwidth]{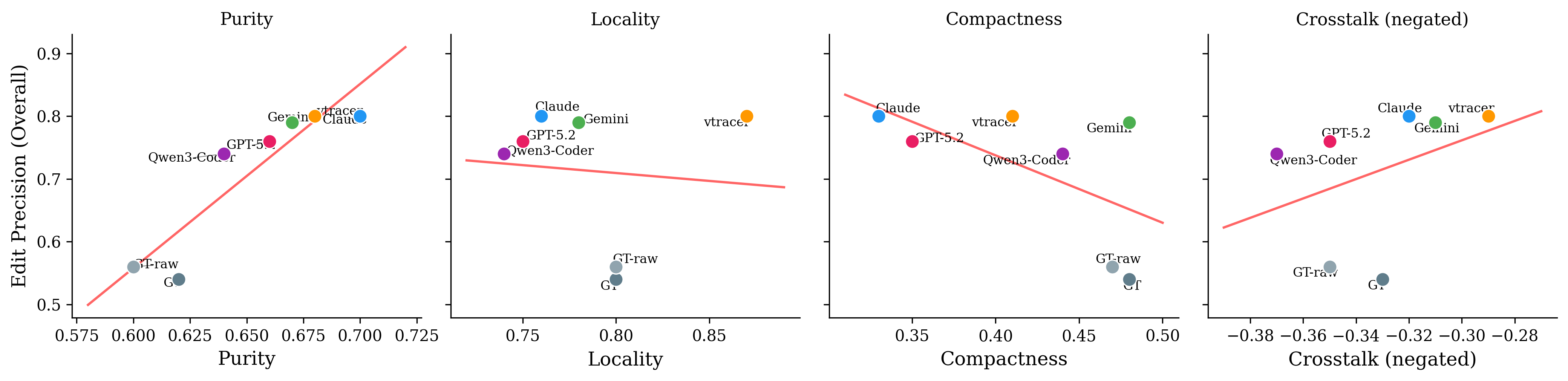}
\caption{Model-level view: mean structural metric vs.\ mean edit precision
(complex tier, 6~models).  Purity visually separates LLM-generated SVGs
(upper right) from source SVGs (lower left); other metrics show weaker
trends.}
\label{fig:model_scatter}
\end{figure}

\paragraph{Metric precision correlation.}
We validate each metric as a predictor of edit precision at two
granularities.  \Cref{fig:binned} shows SVG-level results: we group SVGs
into purity quintiles and plot mean edit precision per bin for each edit
type.  Purity shows a consistent monotonic trend across all five edit types
($r \geq +0.29$, $p < 0.001$ for color/delete/move/scale;
$r = +0.10$, $p = 0.02$ for regroup), with each quintile step
corresponding to roughly 4--5 percentage points of edit precision.
\Cref{fig:model_scatter} provides a qualitative model-level view: purity
cleanly separates LLM-generated SVGs from source SVGs, while other
metrics show weaker trends.

To guard against overfitting to a single tier, we verify that the
correlation holds across complexity levels: purity computed on the complex
tier predicts edit precision on the medium tier ($r = +0.11$, $p < 0.01$,
Bonferroni-corrected), and vice versa ($r = +0.09$).  All four metrics
remain significant in both directions.

\paragraph{Why does Source SVG score lowest?}
Source SVGs have high coverage (0.98) but the lowest purity (0.60)
and edit precision (0.56).  Inspection reveals that 98\% of complex-tier
source SVGs consist exclusively of \texttt{<path>} elements with no
\texttt{<rect>}, \texttt{<circle>}, or \texttt{<text>}; they are
auto-exported files where separate concepts have been merged into compound
paths for file-size reduction.  The entanglement is inherent to the
original files, not introduced by post-processing.  In contrast,
LLM-generated SVGs use separate elements per concept because this is how
SVG is written in training data.

\section{Discussion}
\label{sec:discussion}

\paragraph{Multiple Editability Metrics.}
We present four different metrics that captures distinct structural properties. Models rank differently on each axis (Claude leads purity; vtracer leads locality), and a single number would obscure these
distinctions.  

\paragraph{Effect sizes.}
Per-edit correlations are modest ($r \leq 0.11$), consistent
with the inherent noise in pixel-based edit measurement and CLIP as
a proxy.  However, the \emph{model-level} edit precision gap is large
(0.80 vs.\ 0.56 for Claude vs.\ Source SVG), and the structural metrics
correctly predict this ranking.  The metrics are most useful for
comparing generators, not for predicting individual edit outcomes.

\paragraph{Applications.}
The metric suite provides a structural axis complementing visual fidelity:
models producing identical renders can be distinguished by code
quality.  LOO scoring also enables training-free artifact removal and
denser reward signals for RL ($N$ per-element signals vs.\ one final score).

\paragraph{Limitations.}
LOO requires $N$ leave-one-out renders plus one complete render per SVG
(minutes for SVGs with hundreds of subpaths).  CLIP similarity is an imperfect proxy for visual quality.
The artifact evaluation uses synthetic injections, though the element
rejection experiment validates on real model outputs.  The concept
extraction pipeline (VLM + CLIPSeg) is nondeterministic and may miss
abstract concepts.  Finally, the edit precision protocol uses the same
concept masks both to define the metrics and to measure edit outcomes; this
shared dependency means the validation is not fully independent, though the
train/test split across complexity tiers provides partial mitigation.

\section{Conclusion}
\label{sec:conclusion}

We introduced LOO element-level analysis for SVG generation, deriving
per-element quality scores, concept--element attribution, and four
structural metrics from a single mechanism.  Purity, coverage, compactness,
and locality characterize SVG modularity from complementary angles, with
purity and locality validated as significant predictors of edit precision.
A consistent finding across 6 models and 3 tiers: LLM-generated SVGs are
structurally more modular than source SVGs, because models produce
one element per concept while source files use entangled compound paths.

\paragraph{Future work.}
An immediate extension is applying the LOO metrics to broader SVG benchmarks whose images decompose cleanly into visual concepts, such as the infographics category of Graphic-Design-Bench~\cite{deganutti2026gdbench} and SVGenius~\cite{chen2025svgenius}.  A second direction is using our element-level LOO scores as \emph{progress rewards} for reinforcement learning.  Recent RL-based SVG generators use holistic image-level rewards, whether pixel and perceptual losses~\cite{rlrf}, hybrid design-aware rewards~\cite{reasonsvg}, progressive curriculum rewards~\cite{svgen}, or multi-task rewards~\cite{wang2026reliable}.  Our per-element LOO signals could provide denser, step-level feedback that penalizes harmful elements during generation rather than after the fact, offering a natural bridge between structural evaluation and RL training.  Beyond SVG, the approach generalizes to other structured generation domains such as HTML, \LaTeX, and CAD, wherever outputs have both a rendered form and a meaningful structural decomposition.

\bibliographystyle{unsrt}
\bibliography{references}

\appendix

\section{Medium and Simple Tier Results}
\label{app:tiers}

\begin{table}[h]
\centering\small
\caption{Structural metrics, medium tier.}
\begin{tabular}{lrrrrrr}
\toprule
Model & Purity & Cover. & Compact. & Locality & Crosstalk & Elem. \\
\midrule
Claude  & 0.61 & 0.97 & 0.38 & 0.75 & 0.36 & 15 \\
Qwen3-Coder & 0.59 & 0.97 & 0.48 & 0.74 & 0.33 & 9 \\
Source SVG & 0.58 & 0.99 & 0.58 & 0.81 & 0.35 & 17 \\
vtracer & 0.58 & 0.98 & 0.42 & 0.78 & 0.41 & 24 \\
Gemini  & 0.56 & 0.97 & 0.44 & 0.74 & 0.38 & 11 \\
GPT-4o  & 0.56 & 0.96 & 0.44 & 0.74 & 0.40 & 18 \\
\bottomrule
\end{tabular}
\end{table}

\begin{table}[h]
\centering\small
\caption{Structural metrics, simple tier.}
\begin{tabular}{lrrrrrr}
\toprule
Model & Purity & Cover. & Compact. & Locality & Crosstalk & Elem. \\
\midrule
vtracer & 0.64 & 1.00 & 0.75 & 0.85 & 0.19 & 4 \\
Qwen3-Coder & 0.60 & 0.94 & 0.80 & 0.87 & 0.20 & 6 \\
Claude  & 0.56 & 0.99 & 0.70 & 0.79 & 0.33 & 5 \\
Gemini  & 0.53 & 0.97 & 0.80 & 0.84 & 0.29 & 3 \\
GPT-4o  & 0.52 & 0.96 & 0.79 & 0.87 & 0.30 & 4 \\
Source SVG & 0.49 & 0.98 & 0.79 & 0.85 & 0.34 & 2 \\
\bottomrule
\end{tabular}
\end{table}

\begin{table}[h]
\centering\small
\caption{Empirical edit precision, medium tier.}
\begin{tabular}{lrrrrrr}
\toprule
Model & Overall & Color & Delete & Move & Scale & Regroup \\
\midrule
Claude  & 0.77 & 0.79 & 0.86 & 0.81 & 0.83 & 0.55 \\
Gemini  & 0.76 & 0.75 & 0.86 & 0.80 & 0.82 & 0.54 \\
Qwen3-Coder & 0.75 & 0.74 & 0.85 & 0.80 & 0.82 & 0.53 \\
vtracer & 0.75 & 0.81 & 0.83 & 0.74 & 0.78 & 0.57 \\
GPT-4o  & 0.72 & 0.72 & 0.81 & 0.74 & 0.77 & 0.57 \\
Source SVG & 0.60 & 0.64 & 0.64 & 0.60 & 0.62 & 0.51 \\
\bottomrule
\end{tabular}
\end{table}

\begin{table}[h]
\centering\small
\caption{Empirical edit precision, simple tier.}
\begin{tabular}{lrrrrrr}
\toprule
Model & Overall & Color & Delete & Move & Scale & Regroup \\
\midrule
Claude  & 0.75 & 0.77 & 0.81 & 0.76 & 0.79 & 0.63 \\
Gemini  & 0.74 & 0.76 & 0.79 & 0.72 & 0.79 & 0.63 \\
Qwen3-Coder & 0.73 & 0.73 & 0.82 & 0.66 & 0.74 & 0.69 \\
GPT-4o  & 0.72 & 0.73 & 0.77 & 0.66 & 0.73 & 0.70 \\
vtracer & 0.71 & 0.77 & 0.74 & 0.63 & 0.72 & 0.69 \\
Source SVG & 0.65 & 0.66 & 0.68 & 0.61 & 0.64 & 0.66 \\
\bottomrule
\end{tabular}
\end{table}

\section{Edit Precision Visualization}
\label{app:edit_precision}

\begin{figure}[h]
\centering
\includegraphics[width=\textwidth]{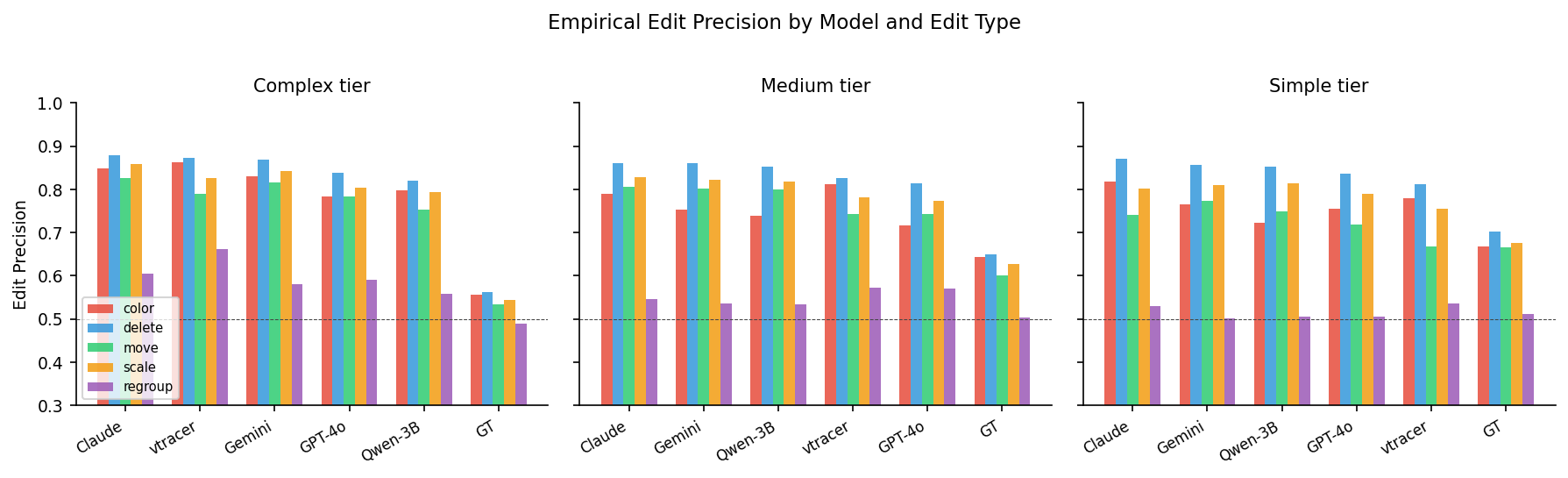}
\caption{Edit precision by edit type across all three tiers.  Delete
consistently achieves the highest precision; regroup the lowest.}
\end{figure}

\end{document}